\documentclass[sigconf, noacm]{acmart}
\usepackage{hhline}
\usepackage{array}
\usepackage{float}
\usepackage{caption}
\usepackage{subcaption}
\usepackage{epstopdf}
\usepackage{gensymb}
\usepackage[normalem]{ulem}
\usepackage{multirow}
\usepackage{multicol}

\newcommand{\etal}{\textit{et al}. }

\AtBeginDocument{%
  \providecommand\BibTeX{{%
    \normalfont B\kern-0.5em{\scshape i\kern-0.25em b}\kern-0.8em\TeX}}}

\acmBooktitle{Proceedings of the 30th ACM International Conference on Multimedia (MM '22), October 10--14, 2022, Lisboa, Portugal}

\acmDOI{10.1145/3503161.3548373}
\setcopyright{none}
\acmSubmissionID{mmfp2871}
\begin{document}

\title{Learning from Label Relationships in Human Affect}

\author{Niki Maria Foteinopoulou}
\affiliation{%
  \institution{Queen Mary University of London}
  \city{London}
  \country{United Kingdom}}
\email{n.m.foteinopoulou@qmul.ac.uk}

\author{Ioannis Patras}
\affiliation{%
  \institution{Queen Mary University of London}
  \city{London}
  \country{United Kingdom}}
\email{i.patras@qmul.ac.uk}

\begin{abstract}
Human affect and mental state estimation in an automated manner, face a number of difficulties, including learning from labels with poor or no temporal resolution, learning from few datasets with little data (often due to confidentiality constraints) and, (very) long, in-the-wild videos. For these reasons, deep learning methodologies tend to overfit, that is, arrive at latent representations with poor generalisation performance on the final regression task. To overcome this, in this work, we introduce two complementary contributions. First, we introduce a novel relational loss for multilabel regression and ordinal problems that regularises learning and leads to better generalisation. The proposed loss uses label vector inter-relational information to learn better latent representations by aligning batch label distances to the distances in the latent feature space. Second, we utilise a two-stage attention architecture that estimates a target for each clip by using features from the neighbouring clips as temporal context. We evaluate the proposed methodology on both continuous affect and schizophrenia severity estimation problems, as there are methodological and contextual parallels between the two. Experimental results demonstrate that the proposed methodology outperforms the baselines that are trained using the supervised regression loss, as well as pre-training the network architecture with an unsupervised contrastive loss. In the domain of schizophrenia, the proposed methodology outperforms previous state-of-the-art by a large margin, achieving a PCC of up to 78\% performance close to that of human experts (85\%) and much higher than previous works (uplift of up to 40\%). In the case of affect recognition, we outperform previous vision-based methods in terms of CCC on both the OMG and the AMIGOS datasets. Specifically for AMIGOS, we outperform previous SoTA CCC for both arousal and valence by 9\% and 13\% respectively, and in the OMG dataset we outperform previous vision works by up to 5\% for both arousal and valence.
\end{abstract}



\renewcommand\footnotetextcopyrightpermission[1]{} 
\pagestyle{plain}
\maketitle

\begin{figure}[!t]
    \includegraphics[width=\linewidth]{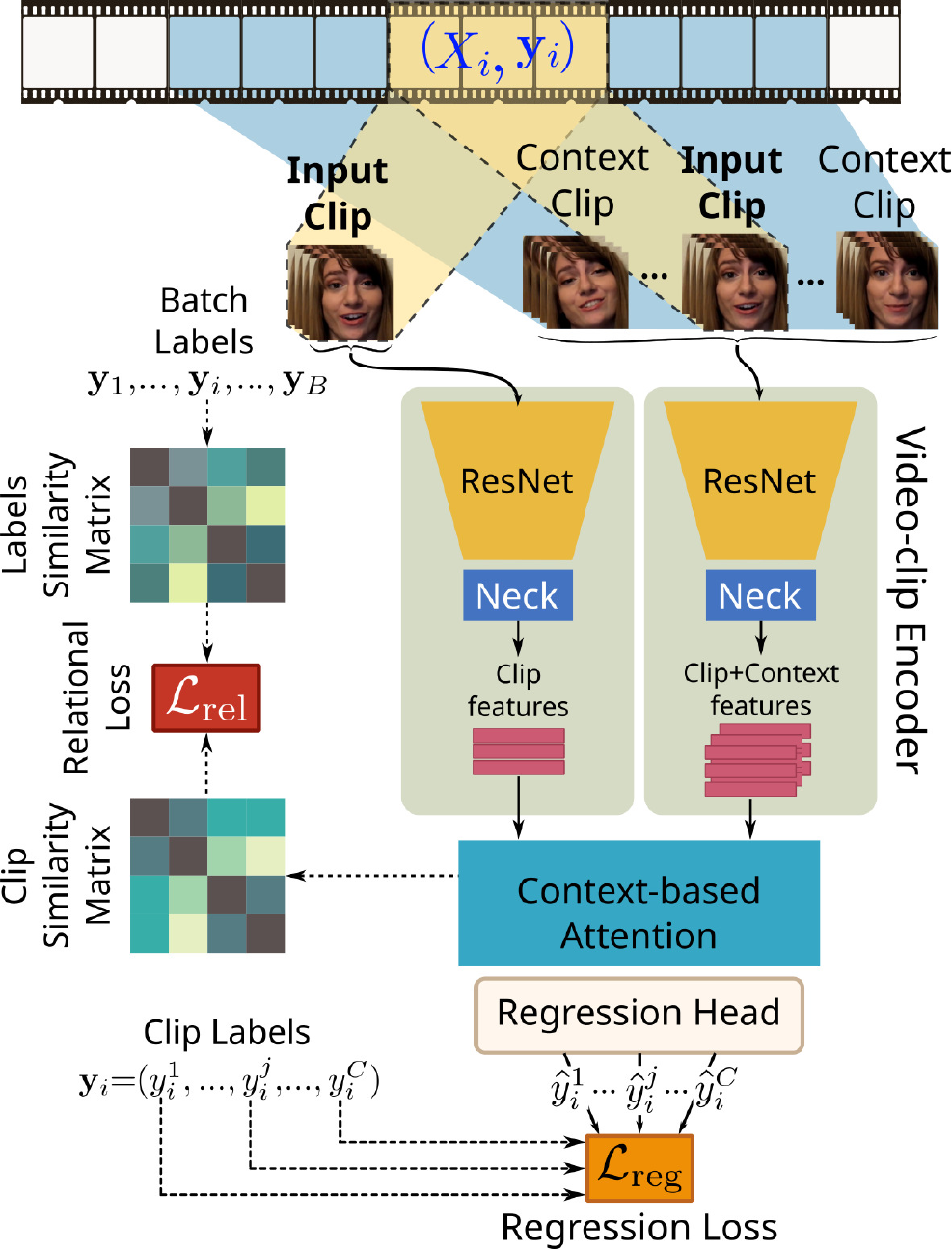}
    \caption{Overview of the proposed framework.}
    \label{fig:summary}
\end{figure}

\section{Introduction}

Understanding human affect and mental state is an active research area with multiple potential applications spanning fields such as education~\cite{YADEGARIDEHKORDI2019103649}, healthcare~\cite{Smith2020}, and entertainment~\cite{games_yannakakis,SETIONO2021781}. For example, by understanding human emotion, the user experience can be enhanced and healthcare professionals can more effectively monitor the patients' emotional state. These problems can be treated either as a classification, using the basic human emotions~\cite{ekman_facial_1978} or by utilising continuous labels along the Arousal-Valence axes~\cite{russell_circumplex_1980}. Similarly, in the domain of mental illness, several scales have been used by healthcare professionals to assess the severity of the symptoms, thus treating symptoms as a spectrum~\cite{american_psychiatric_association_diagnostic_2013}. 

Regardless of which of the above labelling approaches is adopted, certain issues render the problem of human affect and mental state estimation challenging. Specifically, in-the-wild datasets tend to include long videos with low or no temporal label resolution -- i.e., a set of labels describes the entire video. This typically occurs as affect and mental health symptom labels refer to abstract behaviour that is not easily captured and is not always objectively defined. The length of the video poses a major difficulty for Machine Learning methods, due to GPU memory constraints. To address this issue, two main approaches are employed in the literature, namely, a) estimating sub-segments of the long videos~\cite{carneiro_de_melo_deep_2020, lu_multiplespatiotemporal_2018} and b) pre-computing features\cite{zhang_spatiotemporal_2020, yang_posebased_2021, bishay_automatic_2020, mou_alone_2019}. For example, in MIMAMO~\cite{deng_mimamo_2020} and the work of Peng \etal~\cite{peng2018deep} a small number of frames is sampled from each clip. However, this disregards information from the remaining video and the clip context. Moreover, as affect and mental state descriptions often refer to a larger context, short clips might not be representative samples. Similarly, estimating per-frame predictions~\cite{melo_depression_2019} disregards clip information and is also suffering from the lack of temporal information. Previous state-of-the-art works in symptom severity estimation~\cite{bishay_schinet_2018}, used statistical representations, such as Gaussian Mixture Models, on a set of per-frame extracted features. However, this approach does not learn from the temporal relationships of frame features. It also does not allow for end-to-end training, therefore does not allow for feature optimisation on the specific task. In order to exploit contextual information and improve clip-level features, Wu et al.~\cite{wu_long-term_2019} proposed the use of Long-Term Feature Banks for the problem of action recognition in videos. However, Long-Term Feature Banks~\cite{wu_long-term_2019} rely on a pre-computed set of features for the context, that does not improve in quality during training. By contrast, in this work, we build upon~\cite{wu_long-term_2019} and use a context feature extractor that updates context features at each iteration, allowing for dynamically computing context features of random clips sampled from a longer video in an end-to-end manner, leading to much shorter training times.

Publicly available datasets for affect and mental health analysis are typically small, which often results in overfitting problems during training. As such, methods that lead to better representations with a small number of samples are paramount to the success of the final regression task. However, several recent works~\cite{khosla_supervised_2020, chen2020big, hirata_making_2021, bulat_pre-training_2021} require pre-training (whether supervised or unsupervised) with very large datasets to achieve better representations before fine-tuning on the final task.  In continuous affect estimation, Kim \etal~\cite{kim_contrastive_2021} binarised labels and used an adversarial loss on the latent feature space, however this approach ignores the continuous nature of Arousal/Valence dimensions.

In order to both alleviate the challenges due to long video input and to improve the feature representations so as to address the multi-label regression problems that arise in the domain of affect and mental health analysis, in this work we propose a) a novel attention-based video-clip encoder that builds upon~\cite{wu_long-term_2019} and utilises the temporal dimension of the input clips and arrives at clip-level predictions that benefit from context clip information, and b) a novel relational regression loss function that aligns the distances in the latent clip-level representations/features to the distances of the labels of the clips in question. An overview of the proposed framework is shown in Fig.~\ref{fig:summary}. Specifically, we propose to jointly train two network branches: a) one that uses the proposed video-clip encoder to extract clip-level features from the input video clips and a set of temporally neighbouring clips, which subsequently feed a regression head in order to infer the desired values and calculate the regression loss, and b) one that uses the proposed video-clip encoder to extract clip-level features from the input video clips, which subsequently feeds the regression head and are further used to construct the intra-batch similarity matrix for calculating the proposed relational loss. To the best of our knowledge, this is the first work that uses label relationships to improve feature representation learning. The proposed regression head employs an attention-based mechanism for fusing clip-level and context features and regressing to the desired continuous values. The main contributions of this work can be summarised as follows:
\begin{itemize}
    \item We build on~\cite{wu_long-term_2019} and propose a two-stage attention architecture that uses features from the clips' neighbourhood to introduce context information in the feature extraction. The architecture is novel in the domain of affect and mental state analysis and, unlike~\cite{wu_long-term_2019}, it does not train a separate model to compute context features, but rather updates its weights during training -- this leads to smaller training times.
    \item We introduce a novel loss, named relational regression loss, that aims at learning from the label relationships of the samples during training. This loss is using the distance between label vectors to learn intra-batch latent representation similarities in a supervised manner. We show in the ablation studies that the improved latent representations obtained with the addition of the relational loss lead to improved regression output, without the use of large datasets.
    \item We show that the methodology achieves results comparable to the state-of-the-art. Specifically, for symptom severity estimation of schizophrenia, our methodology outperforms the previous state of the art on all scales and symptoms tested and achieves a Pearson's Correlation Coefficient similar to that of human experts.
\end{itemize}

\section{Related Work}

In this Section, we review previous works on continuous affect estimation and mental health assessment and focus on representation learning and temporal context exploitation in large videos.

\subsection{Learning Representations and Label Relations}
In human affect problems and even more in mental state estimation, learning features representative of the behaviour rather than other entangled factors (eg. identity) is paramount to the reliability of the final estimate.
 A number of works have addressed the issue of representation learning, with more recent developments in contrastive methodologies~\cite{chen2020big, chen2020simple}, whether evaluating results on static image data or video datasets~\cite{qian_spatiotemporal_2021}. These self-supervised methodologies learn latent representations by teaching the architecture which data points are similar. By extending the idea of comparing samples, supervised contrastive frameworks propose that images~\cite{khosla_supervised_2020} or videos~\cite{hirata_making_2021} from the same class are treated as similar, which results in embeddings from the same class being more closely aligned. However, these works are trained on very large datasets which are not typically available in affective and mental health problems, and have only been evaluated on classification problems. Kim \etal~\cite{kim_contrastive_2021} implement an adversarial loss to learn better representations for continuous affect, however, Arousal/Valence values are binarised for the adversarial task. In our work, we explore the idea of learning representations by comparing sample similarities in a supervised approach, however we implement a non-binary approach which is more suitable for multi-label regression problems.

Several problems/datasets in the field of continuous affect and mental health have multiple labels, in order to describe various affective attributes/ psychological symptoms. Treating each label independently~\cite{deng_mimamo_2020} ignores their potential correlations as well as increases training times significantly with each additional label.
Several works investigate multi-label recognition problems using graph learning approaches to model label correlations and co-occurrences~\cite{wang_multi-label_2020, LI202253}. However, such approaches do not learn from label similarities \textit{between} samples and do not project these similarities to the latent representation space. In contrast, our work uses information from the inter-sample label similarities to learn better latent representations. 

\begin{figure*}[t]
        \includegraphics[width=\linewidth]{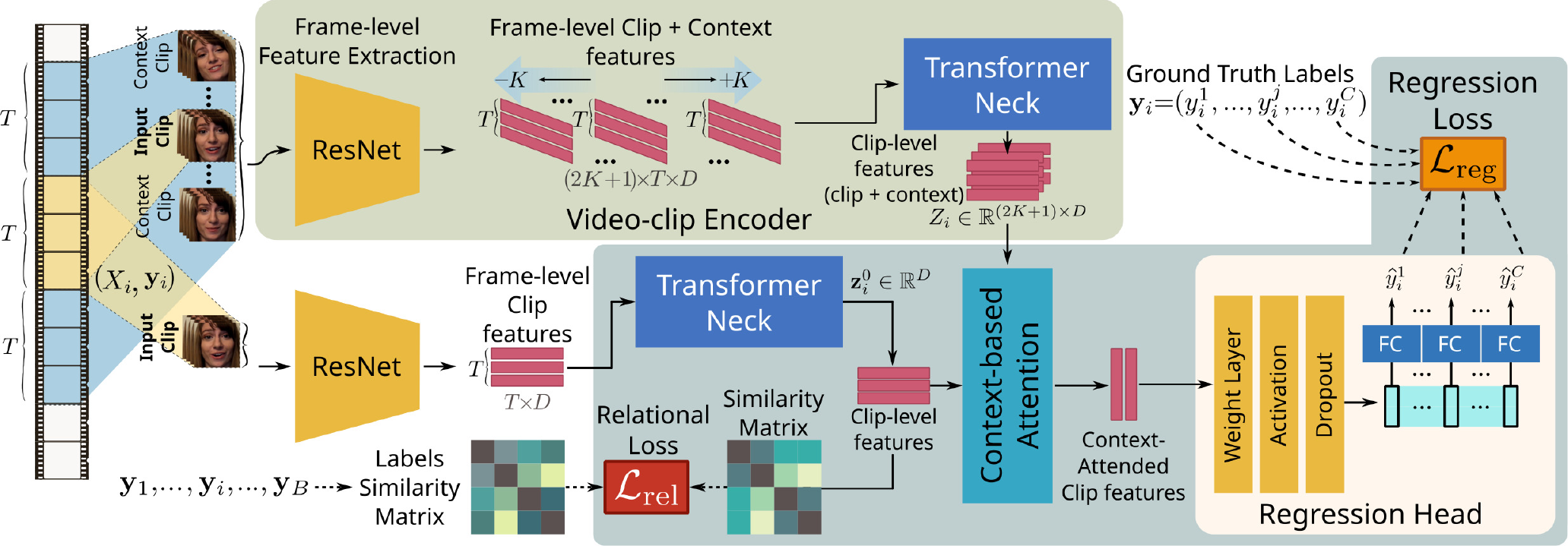}
        \caption{Overview of the proposed framework: (a) The \textit{bottom branch} uses the proposed video-clip encoder (comprising of a ResNet frame-level and a Transformer clip-level feature extractors) to extract clip-level features from the input video clips, which subsequently feed the context-based attention block and are further used to construct the intra-batch similarity matrix for calculating the proposed relational loss. (b) The \textit{upper branch} uses the proposed video-clip encoder to extract clip-level features from the input video clips and a set of context clips from each of the input clips, which subsequently feed the context-based attention block in order to infer the desired values and calculate the regression loss. The context-based attention block, fuses clip-level and context features and passes the context attended clip features to the regression head that estimates the desired continuous values. Error is back-propagated only through the shaded region of the bottom branch.}
        \label{fig:overview}
    \end{figure*}

\subsection{Addressing large sequences}
The exploration of methods tailored to long-range video understanding, is vital for human affect and mental state estimation, as long videos are typically more representative of real-life settings. Moreover, long temporal relationships intuitively should contribute to more accurate estimates of human affect and mental states.
To address long video sequences, previous works have used a number of strategies. One such method is to pre-compute features~\cite{li_temporal_2017, bishay_schinet_2018}; this however does not allow for end-to-end training and makes augmentation techniques more complicated (if feasible at all). Another strategy to address long video sequences is by using contextual features either in the form of intra-sample relations~\cite{zhou_graph-based_2021} or by exploring feature banks~\cite{wu_long-term_2019}. Both of these approaches utilise relations between the short term actions, which is the temporal context of a clip. However, these methods have been evaluated on action recognition problems and have not been implemented in affect. Moreover, while action problems benefit from from long-term context, they still have a much lower label resolution. Finally, in~\cite{wu_long-term_2019} context features need to be pre-computed on pre-defined clips, therefore feature quality does not improve with further training.

In our work, we build on the concept of exploiting contextual features, however differently from~\cite{wu_long-term_2019} we use context features to improve clip-level prediction with end-to-end training. We also do not operate on pre-defined clips, but rather dynamically compute features from them -- with this approach as training progresses, the network learns from better context features.

\subsection{Affect and Mental Health}
As several mental health illnesses and disorders have non-verbal behaviour symptoms, understanding patients' affect is important in diagnosis and severity estimation.
 Depression is a mood disorder that has an impact on patients' affective state; similarly, a number of schizophrenia negative symptoms refer to patients' affect and expressions, therefore there are important semantic parallels between continuous affect estimation and mental health assessment, so we examine them in parallel. More work has been performed on estimating depression severity than symptoms of schizophrenia, as there are no publicly available datasets for the latter.

A number of previous works use a bag of words approach for gestures and facial expressions~\cite{RN14,RN6} or use statistical representations of pre-computed features~\cite{bishay_automatic_2020,tran_modeling_2021,foteinopoulou_estimating_2021}. Some of the symptoms have a quantitative measure (e.g., reduced gestures), therefore, intuitively, methodologies that implicitly measure quantity of features have achieved state-of-the-art results in these previous works. However, such methods disregard the temporal relationship of features. Similarly in~\cite{melo_depression_2019}, by making per-frame predictions for depression estimation, the temporal dimension is not taken into consideration. More recently, in~\cite{huang_assessing_2022} the temporal dimension is taken into consideration as the work leverages audio and text modalities, however, no modality for vision is implemented. In contrast, our work proposes a transformer-based architecture to learn from the temporal dimension. We implement this on RGB frames cropped to the subjects' faces, rather than a set of pre-computed features such as Facial Action Units~\cite{ekman_facial_1978}.

\section{Proposed method}\label{sec:method}
    
    An overview of the proposed framework for the problem of multi-label regression from a sequence of clips is given in Fig.~\ref{fig:overview}. In a nutshell, the proposed architecture consists of two branches with shared weights, that incorporate two main components: a) a video-clip encoder employing a convolutional backbone network for frame-level feature extraction and b) a Transformer-based network leveraging the temporal relationships of the spatial features for clip-level feature extraction (Sect.~\ref{sec:video_clip_encoder}). The clip and context features produced by the aforementioned branches are passed to a context-based attention block (Sect.~\ref{sec:context_attention}) and a regression head (Sect.~\ref{sec:regression_head}). The proposed method uses the context-based attention block to incorporate features from the two branches before passing them to the regression head, as shown in Fig.~\ref{fig:overview}. The bottom branch uses the proposed video-clip encoder to extract clip-level features from the input video clips, which subsequently feed the context-based attention block and are further used to construct the intra-batch similarity matrix for calculating the proposed relational loss (Sect.~\ref{sec:relational_loss}). The goal of the proposed relational loss, as an additional auxiliary task to the main regression, is to obtain a more discriminative set of latent clip-level features, by aligning the label distances in the mini-batch to the latent feature distances. Finally, the upper branch uses the proposed video-clip encoder to extract clip-level features from the input video clips and a set of context clips from each of the input clips, which subsequently feed the regression head in order to infer the desired values and calculate the regression loss.
    
    \subsection{Video-clip encoder}\label{sec:video_clip_encoder}
        
        Let $\mathcal{X}$ be a batch of labelled clips designed so as it contains consecutive clips taken from different video sequences; i.e., $\mathcal{X}=\{(X_i,\mathbf{y}_i)\}_{i=1}^{B}$, where $X_i\in\mathbb{R}^{T\times H\times W\times3}$ denotes the $i$-th clip in the mini-batch, $T$ denotes its duration in frames, $H,W$ denote the frame height and width, $\mathbf{y}_i=(y_1,\ldots,y_C)\in\mathbb{R}^C$ denotes the corresponding ground truth label vector with continuous annotation for $C$ classes, and $B$ denotes the mini-batch size.
        
        Given an input clip $X_i$, the proposed video-clip encoder extracts frame-level features by feeding them to a backbone convolutional network (e.g., a ResNet~\cite{he_deep_2016}), which subsequently feeds a Transformer-based network for extracting clip-level features, leveraging this way the temporal relationships of the calculated spatial features. In the proposed framework, we use the above video-clip encoder in both branches as shown in Fig.~\ref{fig:overview} -- i.e., for calculating the clip-level features $\mathbf{z}_i^0\in\mathbb{R}^D$ for the input clips $X_i$, $i=1,\ldots,B$ (bottom branch) and for calculating clip-level features $Z_i=\left(\mathbf{z}_i^{-K},\ldots,\mathbf{z}_i^{0},\ldots\mathbf{z}_i^{K}\right)\in\mathbb{R}^{(2K+1)\times D}$ from each $X_i$ along with a number $K$ of context clips before and after it (upper branch). 

    \subsection{Context-based Attention}\label{sec:context_attention}
        As discussed above, for any given clip $X_i$ and $2K$ context clips around it, the proposed video-clip encoders extract the clip-level features $\mathbf{z}_i^0\in\mathbb{R}^D$ (corresponding to the input clip $X_i$ alone) and $Z_i=\left(\mathbf{z}_i^{-K},\ldots,\mathbf{z}_i^{0},\ldots\mathbf{z}_i^{K}\right)\in\mathbb{R}^{(2K+1)\times D}$ (corresponding to the input clip $X_i$ and $K$ clips before and $K$ clips after it). These features are then fed to the regression head (Fig.~\ref{fig:overview}), where they are first passed through an attention module before being concatenated. The resulting context-attended clip features are passed to the regression head for the final regression task.
        
    \subsection{Multi-label regression head}\label{sec:regression_head}
        The context-attended clip features obtained through staged attention as explained in the previous sections, is passed through an MLP regression head that predicts the regression values $\hat{\mathbf{y}}_i=(\hat{y}_i^1,\ldots,\hat{y}_i^C)$, $i=1,\ldots,C$. Finally, we calculate the regression loss $\mathcal{L}_{\text{reg}}$ by either using the Root Mean Square Error (RMSE) or the Concordance Correlation Coefficient (CCC), depending on the task at hand, as we will discuss in Sect.~\ref{sec:experimental_setup}.
    
    \subsection{Relational loss}\label{sec:relational_loss}
        At each training iteration, after having calculated (as discussed in Sect.~\ref{sec:video_clip_encoder}) the clip-level features for the clips in a mini-batch, i.e., $\mathbf{z}_i^0\in\mathbb{R}^D$, $i=1,\ldots,B$, we calculate the proposed relational loss as follows:
        \begin{equation}\label{eq:rel_loss}
            \mathcal{L}_{\text{rel}} = \sqrt{\frac{1}{B^2} \sum_{i=1}^B \sum_{j=1}^B \left(\hat{M}_{i,j}-M_{i,j}\right)^2},
        \end{equation}
        where 
        $\hat{M}\in\mathbb{R}^{B\times B}$ denotes the cosine similarity matrix calculated on the clip-level features, whose $(i,j)$-th element is given as 
        $$
            \hat{M}_{i,j}=\frac{\mathbf{z}_i^0\cdot\mathbf{z}_j^0}{\lVert\mathbf{z}_i^0\rVert\lVert\mathbf{z}_j^0\rVert},
        $$
        and $M\in\mathbb{R}^{B\times B}$ denotes the cosine similarity matrix calculated on the ground truth labels, whose $(i,j)$-th element is given as 
        $$
            M_{i,j}=\frac{\mathbf{y}_i\cdot\mathbf{y}_j}{\lVert\mathbf{y}_i\rVert\lVert\mathbf{y}_j\rVert}.
        $$
        
        It is worth noting that, for the calculation of the proposed relational loss, we use the clip-level features from the given clips without using any context clips, in contrast to the regression loss where additional context clips are being used, as discussed in Sect.~\ref{sec:regression_head}. The total loss is then calculated as $\mathcal{L}_{\text{total}}=\mathcal{L}_{\text{reg}} + \lambda\mathcal{L}_{\text{rel}}$, where $\lambda$ is a weighting hyper-parameter which we discuss in Sect.~\ref{sec:experimental_setup}.

    \subsection{Implementation details}\label{sec:implementation_details}
        
        \subsubsection{Backbone frame-level feature extractor}
            We use a standard ResNet50~\cite{he_deep_2016} pre-trained on VGGFace2~\cite{cao2018vggface2} and fine-tuned on FER2013~\cite{goodfellow_challenges_2013} as described in~\cite{albanie_learning_2016}. The classification layer of the pre-trained network was replaced with a fully connected (FC) layer that was fine-tuned for our task during the training of the network, followed by a ReLU~\cite{glorot2011deep} activation. The adopted backbone network receives an input of shape $H \times W \times 3$, where $H,W$ are the height and width of the input frame, respectively, and are set to 224 pixels, and outputs a feature vector with 2048 dimensions for each frame. The per-frame feature vectors are stacked to a matrix of size $T\times2048$ for each clip, where $T$ is the number of frames of each input clip.
            
        \subsubsection{Transformer neck clip-level feature extractor}
            A transformer encoder architecture is employed to learn from the temporal relationships of the spatial feature vectors calculated by the convolutional frame-level feature extractor. The $T\times2048$ features are positionally encoded and fed forward to a Transformer Encoder~\cite{vaswani_attention_2017}. An element-wise addition is performed between the transformer encoder output and the frame-level features, followed by an average pooling operation along the temporal dimension, resulting in a $D$-dimensional clip-level representation, where $D=2048$.
        
        \subsubsection{Context-base Attention}
            For each input clip $X_i$, the regression head takes as input both the clip-level features $\mathbf{z}_i^0\in\mathbb{R}^D$ and the stacked context features $Z_i=\left(\mathbf{z}_i^{-K},\ldots,\mathbf{z}_i^{0},\ldots\mathbf{z}_i^{K}\right)\in\mathbb{R}^{(2K+1)\times D}$ (Sect.~\ref{sec:video_clip_encoder}). A modified non-local block~\cite{wu_long-term_2019} is then used as an attention operation, where clip-level features $\mathbf{z}_i^0$ are used as the query values to attend to features in $Z_i$, which are used as keys and values. The output context attention vector is concatenated with the clip-level features, resulting in a $2\times D$ dimensional vector.
            
        \subsubsection{Regression head}
            The penultimate feature vector is obtained by passing the context-attended feature vector through an FC layer followed by a ReLU activation and a dropout layer.
            
            Finally, in order to obtain the final regression predictions, we split the aforementioned penultimate feature vector into $C$ subsets and attach an FC layer to each subset to obtain the final regression predictions. In the case of continuous affect estimation, we set $C=2$ (i.e., for Arousal/Valence estimation), while for the schizophrenia symptom severity estimation, we set $C$ accordingly to the number of symptoms provided by the scale at hand. Specifically, the CAINS-EXP scale has 4 symptoms in total therefore we set $C=4$. The PANSS Negative scale has 7 symptoms in total, however we select 3 for comparison with previous works~\cite{bishay_schinet_2018, tron_facial_2016}. As the PANSS-NEG scale includes a number of symptoms we do not consider, we add an additional subset in the penultimate feature vector so that $C=4$, which is only considered in the total score estimation. We note that, in the case of symptom severity estimation, additionally to each individual symptom prediction, we predict a total score (by using an additional FC layer) using the entire aforementioned penultimate feature vector. This is in contrast to~\cite{bishay_schinet_2018}, where the total score is estimated using the individual symptom scores.

\section{Experimental setup}\label{sec:experimental_setup}

    \subsection{Datasets}
    
        \textbf{NESS:} The dataset was originally collected to study the effect of group body psychotherapy on negative symptoms of schizophrenia~\cite{priebe_effectiveness_2016}. The participants in this study were recruited from mental health services from different parts of the UK. In total 275 participants were interviewed at three different stages of the study: a) a baseline, b) at the end of the treatment, and c) after six months. Each clinical interview recording is between 40-120 minutes long and is performed in-the-wild, reflecting this way the conditions of real-life clinical interviews. Each interview is assessed in terms of two symptom scales, namely, PANSS~\cite{kay_positive_1987} and CAINS~\cite{forbes_initial_2010}. Out of the total 275 patients, 110 accepted to be recorded at baseline, 93 at end of treatment, and 69 in the six months follow up. The videos in the dataset were recorded at various resolutions and frames per second, however we standardised the resolution to $1920\times1080$ and fps to 25 frames/s for all videos and we discarded videos where a face was not detected on more than 10\% of the frames. Training and evaluation were performed on videos recorded at baseline, for a fair comparison with works in the literature, i.e., 113 videos for 69 patients. All results reported on this dataset are based on a leave-one-patient-out cross-validation scheme, where videos were down-sampled to 3 fps. The values for ``Total Negative'' and ``EXP - Total'' in the PANSS and the CAINS scales, respectively, were scaled during training to match the range of individual symptoms (i.e., 1-7 for PANSS and 0-4 for CAINS).
        
        \textbf{OMG:} The ``OMG-Emotion Dataset''~\cite{barros_omg-emotion_2018} consists of in-the-wild videos of recorded monologues and acting auditions, collected from YouTube. Multiple annotators separated each clip into utterances and assigned labels for Arousal in the $[0,1]$ scale and Valence in the $[-1,1]$ scale. The dataset originally consisted of training, validation, and test sets with a total of 7371 utterances. As a number of videos have been removed since the publication of the dataset, we trained on 2071 and evaluated on 1663 utterances. We also scaled Arousal to $[-1, 1]$ to match the range of Valence during training and inference.
        
        \textbf{AMIGOS:} The AMIGOS dataset~\cite{miranda_correa_amigos_2018} consists of audio-visual and physiological responses of participants (either alone or in a group) to a video stimulus. In this work, we used the responses of individuals; i.e., where 40 participants watched 16 short videos and 4 long ones. The former were defined as videos of 50-150 seconds. The responses were broken down to $20$-second intervals and annotated by three annotators for \textit{Arousal} and \textit{Valence} on a $[-1,1]$ scale. We extracted the frames from the video ($6$ frames/s) and calculated the average score of the three annotators as the ground truth during training for the video segment. We trained the network following a leave-one-subject-out cross validation scheme. At each fold we randomly selected a subset of the training data, corresponding to 20\% of samples. This is to show how the relational loss can achieve state-of-the-art results using a much smaller number of samples than conventional supervised methodologies.
        
        \begin{table}[t]
            \centering
            \caption{Performance (CCC) of the proposed method against baseline and other uni-modal architectures (OMG).}
            \label{tbl:omg_lit}
            \begin{tabular}{|l||cc|} 
            \hline
                                                     & Arousal       & Valence        \\ 
            \hhline{|=::==|}
            Proposed                                 & \uline{0.26}  & \textbf{0.48}  \\ 
            Proposed w/o $K$                     & \textbf{0.29} & \uline{0.46}   \\ 
            Proposed w/o $K$ w/o $\mathcal{L}_{rel}$ & 0.24          & 0.44           \\ 
            Proposed w/ $\mathcal{L}_{cont.}$                     & 0.15             & 0.32              \\ 
            \hhline{|=::==|}
            Peng \etal~\cite{peng2018deep}                                   & 0.24          & 0.43           \\ 
            Kollias and Zafeiriou~\cite{kollias_multi-component_2019}                   & 0.13          & 0.40           \\
            \hline
            \end{tabular}
        \end{table}

    \subsection{Augmentation}
        During training, we applied data augmentation to the spatial dimensions of all datasets. Specifically, we randomly changed the contrast, the saturation, and the hue of frames with a factor of 0.2 and we applied random horizontal flipping and random rotations (with a range of $30\degree$). The same set of transformations was applied to all frames within a clip. Moreover, as clips with temporal length $T$ were selected from a larger video, we considered the clipping along the temporal dimension as an augmentation approach. More specifically, from the video sequence, we selected a random initial frame and selected $T$ consecutive frames to form a clip. Similarly, the context clips were defined as clips with $T$ number of frames that were positioned before and after the current clip in the video sequence. If the initial frame selected did not allow us to define a complete clip, we looped the video. The number of frames $T$ was set to 32 for the experiments conducted on the NESS, and to 16 for the experiments conducted on the OMG and the AMIGOS datasets.

    \begin{table}[!t]
    \centering
    \caption{Effect of number of frames $T$ in terms of CCC (OMG).}
    \label{tbl:omg_t}
    \begin{tabular}{|l||lll|} 
    \hline
             & Arousal       & Valence       & Mean            \\ 
    \hhline{|=::===|}
    $T = 8$  & 0.25          & 0.41          & 0.33            \\
    $T = 16$ & \textbf{0.26} & \textbf{0.49} & \textbf{0.375}  \\
    $T = 32$ & 0.19          & 0.40          & 0.295           \\
    \hline
    \end{tabular}
    \end{table}

    \begin{table*}[!t]
    \footnotesize
    \centering
    \caption{Effect of $T$ on the PANSS-NEG symptom scale.}
    \label{tbl:panss_t}
    \begin{tabular}{|l||lll||lll||lll||lll|} 
    \hline
                & \multicolumn{3}{l||}{N3: Poor Rapport}        & \multicolumn{3}{l||}{N6: Lack of Spontaneity} & \multicolumn{3}{l||}{N1: Blunted Affect}      & \multicolumn{3}{l|}{Total Negative}            \\ 
    \cline{2-13}
                & MAE           & RMSE          & PCC           & MAE           & RMSE          & PCC           & MAE           & RMSE          & PCC           & MAE           & RMSE          & PCC            \\ 
    \hhline{|=::===::===::===::===|}
    $T = 8$~  & 0.97          & 1.31          & 0.67          & \textbf{0.61} & \textbf{0.80} & \textbf{0.66} & \textbf{0.62} & 0.88          & 0.55          & 3.27          & 4.14          & 0.64           \\ 
    \hline
    $T = 16$  & 0.98          & 1.29          & 0.70          & 0.68          & 0.86          & 0.62          & 0.68          & 0.93          & 0.45          & 3.59          & 4.54          & 0.54           \\ 
    \hline
    $T= 32$   & \textbf{0.87} & \textbf{1.16} & \textbf{0.78} & 0.74          & 0.95          & 0.47          & 0.64          & \textbf{0.87} & \textbf{0.56} & \textbf{2.80} & \textbf{3.78} & \textbf{0.71}  \\
    \hline
    \end{tabular}
    \end{table*}
    
    \subsection{Training}
        During training, the hyperparameter $\lambda$ that scales the relational loss was empirically set to 2 for experiments conducted on the NESS and the AMIGOS datasets, and to 1 for experiments conducted on the OMG dataset. During testing, the clips were generated by a sliding window over the video sequence, resulting in non-overlapping clips; the average prediction of all clips in the video was calculated to estimate the final predicted label vector. 
        The network was trained in an end-to-end manner with a batch size of 4, 8, and 16 for the NESS, the OMG, and the AMIGOS datasets, respectively, keeping the pre-trained weights of the ResNet-50 backbone frozen. We used an Adam optimizer with an initial learning rate of $10^{-4}$, multiplied by 0.1 every 5 epochs, and weight decay $5\cdot10^{-3}$. The hyperparameter $K$ that controls the context window size was set to 2 for the experiments on the NESS and to 1 for the experiments on the OMG and the AMIGOS datasets. The network incorporated an RMSE loss during training for the experiments conducted on the NESS and (1-CCC) for the experiments on the OMG and AMIGOS datasets, as proposed by previous works in continuous affect~\cite{toisoul_estimation_2021, deng_mimamo_2020}.
    
    \subsection{Architecture Complexity}
        The proposed architecture has ~90M trainable parameters, distributed as 4M in the backbone, 52M in the transformer neck and 33M in the context aware attention and regression head. We note that, even though the architecture is using two branches (one for clip level features and one for context features), the two branches share weights which significantly reduces the number of parameters. We also note, that similarly to other state-of-the-art methods~\cite{deng_mimamo_2020, kollias_multi-component_2019}, we use a ResNet50 as our backbone network, but in contrast to them that employ an RNN architecture to explore the temporal relationships, we instead use a Transformer Encoder module. As shown in~\cite{vaswani_attention_2017}, the self-attention layers of the Transformer are both faster and less complex than recurrent layers (RNN) when the sequence length is shorter than the feature dimensionality, which is the case in the current architecture, hence, the proposed method is more efficient than RNN-based two-stream methods. We report that the inference time is on average at 28.6ms (±2ms) for a clip prediction.

\section{Results and Discussion}\label{sec:experimental_results}

    In this section we present the experimental evaluation of the proposed framework. We begin with our ablation study in Sect.~\ref{sec:ablation_study}, in order to demonstrate the effectiveness of our method with respect to various design options. Then, in Sect.~\ref{sec:comp_to_soa}, we present comparisons with state-of-the-art methods, where we show that the proposed method achieves results comparable to the state-of-the-art -- specifically, for symptom severity estimation of schizophrenia, our method outperforms the previous state-of-the-art on all scales and symptoms tested and achieves a Pearson's Correlation Coefficient similar to that of human experts.
    
    \begin{figure}[!t]
            \centering
            \begin{subfigure}[b]{0.24\textwidth}
            \centering
            \includegraphics[width=\textwidth]{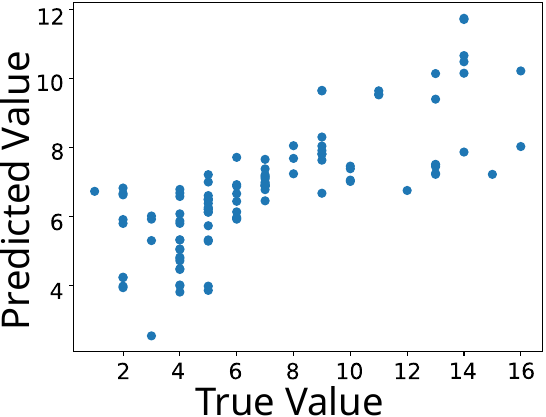}
            \caption{CAINS: EXP -- Total | PCC: 0.77}
            \end{subfigure}
            ~
            \begin{subfigure}[b]{0.25\textwidth}
            \centering
            \includegraphics[width=\textwidth]{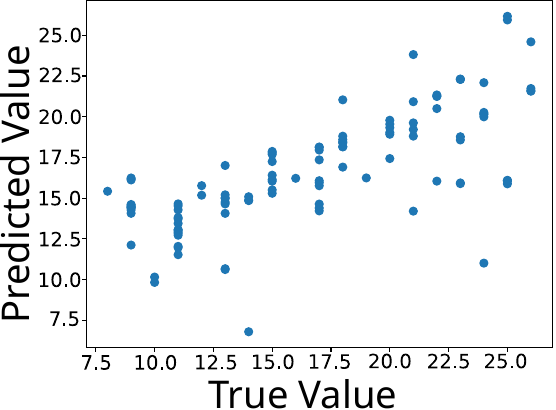}
            \caption{PANSS: NEG -- Total | PCC: 0.71}
            \end{subfigure}
            \caption{Scaled ``Total Score'' estimations of the proposed method on NESS using (a) CAINS and (b) PANSS scales.}
            \label{fig:ness_res}
        \end{figure}
        
        \begin{figure}[!t]
            \includegraphics[width=\linewidth]{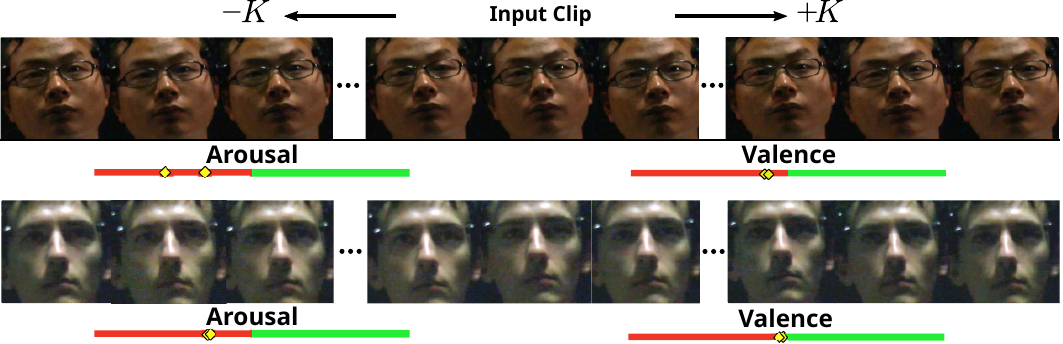}
            \caption{Examples of input clips, their context, and proposed method output on the AMIGOS~\cite{miranda_correa_amigos_2018} dataset: In the top row our method predicted A:$-0.22$, V:$-0.12$ (ground truth: A:$-0.42$, V:$-0.12$), and in the bottom row A:$-0.29$, V:$-0.03$ (ground truth: A:$-0.29$, V:$-0.04$).}
            \label{fig:sample}
        \end{figure}
    
    
    \begin{table*}[!t]
            \footnotesize
            \centering
            \caption{Ablation study on the PANSS-NEG symptom scale.}
            \label{tbl:panss_abl}
            \begin{tabular}{|l||ccc||ccc||ccc||ccc|} 
            \hline
            \multicolumn{1}{|c||}{\multirow{2}{*}{}}                                           & \multicolumn{3}{c||}{N3: Poor Rapport}        & \multicolumn{3}{c||}{N6: Lack of Spontaneity} & \multicolumn{3}{c||}{N1: Blunted Affect}      & \multicolumn{3}{c|}{Total Negative}            \\ 
            \cline{2-13}
            \multicolumn{1}{|c||}{}                                                            & MAE           & RMSE          & PCC           & MAE           & RMSE          & PCC           & MAE           & RMSE          & PCC           & MAE           & RMSE          & PCC            \\ 
            \hhline{|=::===::===::===::===|}
            Proposed                                                                           & \textbf{0.87} & \textbf{1.16} & \textbf{0.78} & \textbf{0.74} & \textbf{0.95} & \textbf{0.47} & \textbf{0.64} & \textbf{0.87} & \textbf{0.56} & \textbf{2.80} & \textbf{3.78} & \textbf{0.71}  \\ 
            \hline
            \begin{tabular}[c]{@{}l@{}}Proposed w/o $\mathcal{L}_{rel}$\end{tabular}             & 1.09          & 1.41          & 0.66          & 0.82          & 0.98          & 0.44          & 0.74          & 0.95          & 0.39          & 3.51          & 4.34          & 0.66           \\ 
            \hline
            \begin{tabular}[c]{@{}l@{}l@{}}Proposed w/o $\mathcal{L}_{rel}$ w/o $K$\end{tabular} & 1.25          & 1.57          & 0.41          & 0.85          & 1.01          & 0.36          & 0.75          & 0.97          & 0.36          & 3.70          & 3.62          & 0.58           \\ 
            \hline
            \begin{tabular}[c]{@{}l@{}}Proposed w/ $\mathcal{L}_{cont.}$\end{tabular}                                                              & 1.09          & 1.53          & 0.41          & 0.88          & 1.05          & 0.19          & 0.77          & 1.01          & 0.35          & 3.56          & 4.48          & 0.55           \\
            \hline
            \end{tabular}
        \end{table*}
            
        \begin{table*}[!t]
            \footnotesize
            \centering
            \caption{Ablation study on the CAINS-EXP symptom scale.}
            \label{tbl:cains_abl}
            \begin{tabular}{|l||lll||lll||lll||lll||lll|} 
            \hline
                                                                                               & \multicolumn{3}{c||}{Facial Expression}                                          & \multicolumn{3}{c||}{Vocal Expression}                                           & \multicolumn{3}{c||}{Expressive Gestures}                                        & \multicolumn{3}{c||}{Quantity of Speech}                                         & \multicolumn{3}{c|}{EXP-Total Score}                                             \\ 
            \cline{2-16}
                                                                                                       & \multicolumn{1}{c}{MAE} & \multicolumn{1}{c}{RMSE} & \multicolumn{1}{c||}{PCC} & \multicolumn{1}{c}{MAE} & \multicolumn{1}{c}{RMSE} & \multicolumn{1}{c||}{PCC} & \multicolumn{1}{c}{MAE} & \multicolumn{1}{c}{RMSE} & \multicolumn{1}{c||}{PCC} & \multicolumn{1}{c}{MAE} & \multicolumn{1}{c}{RMSE} & \multicolumn{1}{c||}{PCC} & \multicolumn{1}{c}{MAE} & \multicolumn{1}{c}{RMSE} & \multicolumn{1}{c|}{PCC}                        \\ 
            \hhline{|=::===::===::===::===::===|}
            Proposed                                                                                   & \textbf{0.56}            & \textbf{0.72}             & \textbf{0.75}             & \textbf{0.65}            & \textbf{0.89}             & \textbf{0.71}             & \textbf{0.71}            & \textbf{0.89}             & \textbf{0.76}             & \textbf{0.60}            & \textbf{0.82}             & \textbf{0.54}             & \textbf{1.88}            & \textbf{2.6}              & \textbf{0.77}                                   \\ 
            \hline
            \begin{tabular}[c]{@{}l@{}}Proposed w/o $\mathcal{L}_{rel}$\end{tabular}                      & 0.59                     & 0.78                      & 0.63                      & 0.75                     & 0.98                      & 0.60                      & 0.72                     & 0.96                      & 0.59                      & 0.62                     & 0.85                      & 0.51                      & 2.12                     & 2.94                      & 0.71 \\ 
            \hline
            \begin{tabular}[c]{@{}l@{}}Proposed w/o $\mathcal{L}_{rel}$ w/o $K$\end{tabular} & 1.06                     & 1.33                      & 0.64                      & 1.06                     & 1.36                      & 0.59                      & 1.14                     & 1.37                      & 0.62                      & 0.77                     & 1.02                      & 0.44                      & 3.76                     & 4.72                      & 0.48                                            \\ 
            \hline
            \begin{tabular}[c]{@{}l@{}}Proposed w/ $\mathcal{L}_{cont.}$\end{tabular}                & 1.12                     & 1.37                      & 0.45                      & 1.06                     & 1.35                      & 0.54                      & 1.19                     & 1.49                      & 0.41                      & 0.84                     & 1.11                      & 0.26                      & 3.87                     & 4.80                       & 0.41                                            \\
            \hline
            \end{tabular}
        \end{table*}
    \subsection{Ablation study}\label{sec:ablation_study}

    In order to examine the effect of number of frames $T$ in the overall method, we train the proposed methodology for $T={8, 16, 32}$ on the OMG and NESS datasets. The results of the ablation on $T$ for the OMG dataset is shown on Table~\ref{tbl:omg_t}; we observe that the highest $CCC$ for both arousal and valence is achieved when $T=16$, closely followed by $T=8$. The effect of $T$ on the PANSS-NEG scale are shown on Table~\ref{tbl:panss_t}; we note that model performance is overall benefited by a larger $T$, with the exception of symptom N6, which is consistent with the symptom definition (i.e., Lack of Spontaneity and Flow in conversation, which is expected to be short-termed).
    
        In order to investigate the effectiveness of the components of the proposed framework, we conducted an ablation study where we gradually excluded the incorporation of contextual clips and the proposed relational loss. For doing so, we trained a baseline network without context features and trained only on the standard regression loss (i.e., without the proposed relational loss), which we denote as ``w/o $K$ w/o $\mathcal{L}_{rel}$''. We also trained a version of the network including the context branch without the relational loss, which we denote as ``w/o $\mathcal{L}_{rel}$''. We finally conducted an experiment using an unsupervised contrastive pre-training, which we denote as ``$\mathcal{L}_{cont.}$''. In this scenario, we firstly pre-trained the clip-level feature extraction backbone in an unsupervised contrastive manner, and then we trained the regression head on top of the frozen backbone, using the regression loss. For the unsupervised contrastive loss, we sampled 2 clips from the same video as positive samples and considered samples from other videos as negatives.
        
        The analysis results on the NESS~\cite{priebe_effectiveness_2016} dataset are shown in Tables~\ref{tbl:panss_abl},~\ref{tbl:cains_abl} for the PANSS and CAINS scales respectively. We see that the proposed network under the contrastive pre-training scenario, has a similar performance to experiments where we trained with only the regression loss (shown as ``w/o $\mathcal{L}_{rel}$'') in terms of MAE/RMSE, however in terms of PCC the non-contrastive network still outperforms the contrastive methodology by a large margin. We attribute this to the size of the dataset that was required to learn discriminative features, as other unsupervised methodologies for representation learning~\cite{chen2020big, chen2020simple, qian_spatiotemporal_2021} trained on very large datasets such as ImageNet~\cite{imagenet_cvpr09} and Kinetics~\cite{kay_kinetics_2017, carreira_short_2018}. Furthermore, the proposed relational clearly leads to a large improvement to the overall regression task, against the baseline and the unsupervised contrastive loss using a small number of training samples. Contextual features also improved the overall regression performance particularly for the MAE/RMSE metrics, with a more noticeable improvement in the Total Scores of the two scales.
         \begin{table*}[!t]
            \footnotesize
            \centering
            \caption{Performance of proposed method against state-of-the-art methods on the PANSS-NEG symptom scale.}
            \label{tbl:panss_lit}
            \begin{tabular}{|l||lll||lll||lll||lll|} 
            \hline
            \begin{tabular}[c]{@{}l@{}}\\\end{tabular} & \multicolumn{3}{l||}{N3: Poor Rapport}                             & \multicolumn{3}{l||}{N6: Lack of Spontaneity}                      & \multicolumn{3}{l||}{N1: Blunted Affect}                           & \multicolumn{3}{l|}{Total Negative}                                 \\ 
            \cline{2-13}
                                                       & MAE                  & RMSE                 & PCC                  & MAE                  & RMSE                 & PCC                  & MAE                  & RMSE                 & PCC                  & MAE                  & RMSE                 & PCC                   \\ 
            \hhline{|=::===::===::===::===|}
            Tron \etal~\cite{tron_automated_2015}                               & 0.98                 & 1.31                 & 0.20                 & 1.37                 & 1.69                 & 0.13                 & 0.90                 & 1.28                 & 0.37                 &          -            &         -             &           -            \\ 
            \hline
            Tron \etal~\cite{tron_facial_2016}                               & 1.01                 & 1.26                 & 0.15                 & 1.32                 & 1.62                 & 0.09                 & 0.99                 & 1.36                 & 0.11                 &          -            &         -             &           -            \\ 
            \hline
            SchiNet~\cite{bishay_schinet_2018}                                   & \textbf{0.85 }       & 1.20         & 0.27                 & 1.25                 & 1.51                 & 0.25                 & 0.84                 & 1.18                 & 0.42                 & 3.30                 & 4.17                 & 0.29                  \\ 
            \hline
            Proposed                                   & \uline{0.87}                 & \textbf{1.16}                 & \textbf{0.78}        & \textbf{0.74}        & \textbf{0.95}        & \textbf{0.47}        & \textbf{0.64}        & \textbf{0.87}        & \textbf{0.56}        & \textbf{2.80}         & \textbf{3.78}        & \textbf{0.71}         \\ 
            \hline
            \end{tabular}
        \end{table*}
    
        \begin{table*}[!th]
            \footnotesize
            \centering
            \caption{Performance of proposed methodology against other state-of-the-art on the CAINS-EXP symptom scale.}
            \label{tbl:cains_lit}
            \begin{tabular}{|l||lll||lll||lll||lll||lll|} 
            \hline
                                 & \multicolumn{3}{l||}{Facial Expression}                      & \multicolumn{3}{l||}{Vocal Expression}                       & \multicolumn{3}{l||}{Expressive Gestures}                    & \multicolumn{3}{l||}{Quantity of Speech}                     & \multicolumn{3}{l|}{EXP-Total Score}                              \\ 
            \cline{2-16}
                                 & MAE                  & RMSE                 & PCC                  & MAE                  & RMSE                 & PCC                  & MAE                  & RMSE                 & PCC                  & MAE                  & RMSE                 & PCC                  & MAE                  & RMSE                 & PCC                   \\ 
            \hhline{|=::===::===::===::===::===|}
            Tron~\etal~\cite{tron_automated_2015}        & 0.80                 & 1.03                 & 0.37                 & 0.87                 & 1.23                 & 0.23                 & 0.85                 & 1.19                 & 0.36                 & 1.09                 & 1.43                 & 0.27                 &        -              &           -           &               -        \\ 
            \hline
            Tron~\etal~\cite{tron_facial_2016}        & 0.75                 & 1.07                 & 0.36                 & 0.86                 & 1.22                 & 0.26                 & 0.91                 & 1.22                 & 0.38                 & 1.02                 & 1.36                 & 0.25                 &           -           &        -              &         -              \\ 
            \hline
            SchiNet~\cite{bishay_schinet_2018}              & 0.66                 & 0.93                 & 0.46                 & 0.77                 & 1.10                 & 0.27                 & 0.90                 & 1.15                 & 0.36                 & 0.98                 & 1.30                 & 0.30                 & 2.67                 & 3.34                 & 0.45                  \\ 
            \hline
            Proposed                                                                                   & \textbf{0.56}            & \textbf{0.72}             & \textbf{0.75}             & \textbf{0.65}            & \textbf{0.89}             & \textbf{0.71}             & \textbf{0.71}            & \textbf{0.89}             & \textbf{0.76}             & \textbf{0.60}            & \textbf{0.82}             & \textbf{0.54}             & \textbf{1.88}            & \textbf{2.60}              & \textbf{0.77}\\
            \hline
            \end{tabular}
        \end{table*}

        \begin{table}[!th]
            \footnotesize
            \centering
            \caption{Performance of the proposed method against baseline and other uni-modal architectures (AMIGOS).}
            \label{tbl:amigos_lit}
            \begin{tabular}{|c||cc||ll|} 
            \hline
            \multirow{2}{*}{}                                                                                                  & \multicolumn{2}{c||}{Arousal} & \multicolumn{2}{l|}{Valence}  \\ 
            \cline{2-5}
                                                                                                                               & PCC    & CCC                  & PCC    & CCC                  \\ 
            \hhline{|=::==::==|}
            Proposed                                                                                                           & \textbf{0.69} & \textbf{0.68}               & \textbf{0.75} & \textbf{0.74}               \\
            \begin{tabular}[c]{@{}l@{}}Proposed $\mathcal{L}_{CCC}$ w/o $K$ w/o$ \mathcal{L}_{rel}$\end{tabular}      & 0.59   & 0.49                 & 0.64   & 0.54                 \\
            \begin{tabular}[c]{@{}l@{}}Proposed~ $\mathcal{L}_{RMSE}$ w/o $K$ w/o $\mathcal{L}_{rel}$\end{tabular} & 0.60   & 0.39                 & 0.55   & 0.40                 \\
            \hhline{|=::==::==|}
            Mou \etal~\cite{mou_alone_2019}                                                                                                             & 0.60   & 0.59                 & 0.62   & 0.61                 \\
            \hline
            \end{tabular}
        \end{table}
        
        The results of our ablation study on the OMG dataset~\cite{barros_omg-emotion_2018} are presented in Table~\ref{tbl:omg_lit}. Comparing the proposed methodology against its baseline (i.e., ``w/o $K$ w/o $\mathcal{L}_{rel}$''), we observe that the proposed relational loss improves the performance of the regression measured in terms of CCC, for both Arousal and Valence. Further incorporating the contextual features improved the CCC score for Valence, but lowered slightly the CCC for Arousal. However, compared to other works submitted to the challenge~\cite{kollias_multi-component_2019,peng2018deep}, the proposed network and specifically the use of the novel relational loss, shows a clear improvement in terms of CCC for both Arousal and Valence. We also observe a clear advantage of the proposed method compared to the architecture pre-trained with contrastive loss, which appears to over-fit and it may be encouraging the network to learn features of the subjects' identities rather than affective and mental states, due to the nature of the problem and database size.

    \subsection{Comparison to state-of-the-art}\label{sec:comp_to_soa}
        In this section we present the results of the proposed method against state-of-the-art methods. The results for the NESS dataset~\cite{priebe_effectiveness_2016} against previous works are shown in Tables~\ref{tbl:panss_lit} and \ref{tbl:cains_lit}, for PANSS and CAINS scales respectively. We can see that the proposed methodology outperforms previous works across all the evaluated symptoms and scales by a large margin, particularly for PCC, achieving state-of-the-art results. Since the NESS dataset has been annotated by different healthcare professionals, we can compare the PCC achieved by the proposed method against the PCC of the annotators (mental health experts), which has a mean value of $\textbf{0.85}$~\cite{bishay_automatic_2020, priebe_effectiveness_2016} on NESS. We observe that the proposed method achieves a PCC close to that of human experts for the ``Total Negative'' and ``EXP-Total'' scores, in this dataset. In Fig.~\ref{fig:ness_res} we show the total score predictions for all videos, for both scales in the NESS dataset. As the NESS dataset is imbalanced, with fewer patients having severe symptoms, we observe a higher error for patients with higher ground truth labels. Moreover, since we perform a leave-one-patient-out cross-validation, there is a chance that no examples of high total scores are included in the training set of a given fold. This trend is consistent for both scales used to evaluate.
        
        For experiments conducted on the OMG dataset~\cite{barros_omg-emotion_2018}, we compared the performance of the proposed method against other uni-modal multi-label works submitted to the ``OMG-Emotion Behavior Challenge'' -- we show the results in Table~\ref{tbl:omg_lit}, where we observe a clear improvement against previous works, for both Arousal and Valence in terms of CCC. We note that, to our knowledge, current state-of-the-art results for the OMG dataset are achieved by MIMAMO~\cite{deng_mimamo_2020} with a CCC of $0.37$ and $0.52$ for Arousal and Valence respectively. However, as MIMAMO is a multi-modal approach (using RGB and inter-frame phase difference as input modalities) and is trained for a single target (i.e., Arousal or Valence) at a time, the results reported in~\cite{deng_mimamo_2020} are not directly comparable to ours.
        
        Finally, for the experiments conducted on the AMIGOS dataset~\cite{miranda_correa_amigos_2018}, we compared the performance of the proposed methodology against previous state-of-the-art~\cite{mou_alone_2019} for the face modality and we show the results in Table~\ref{tbl:amigos_lit}. The proposed methodology leads to a clear improvement against both baselines, trained with an RMSE regression loss ($\mathcal{L}_{RMSE}$) and a CCC loss ($\mathcal{L}_{CCC}$). We also outperform previous state-of-the-art by a large margin for both Arousal and Valence, even though we trained on a subset of the training data at each fold. It is worth noting that on the AMIGOS dataset the architecture that was pre-trained with a contrastive loss completely overfitted on the regression task and, thus, we choose to excluded it from the comparison. In Fig.~\ref{fig:sample}, we see some visual examples of input clips, their context from the AMIGOS dataset~\cite{miranda_correa_amigos_2018} and the proposed methodology predictions against the ground truth.

\section{Conclusion}

    In this work we presented our method for dealing with challenges that arise in the domain of affect and mental health in multi-label regression problems. Specifically, we built on~\cite{wu_long-term_2019} and proposed a two-stage attention architecture that uses features from the clips' neighbourhood to introduce context information in the feature extraction. The architecture is novel in the domain of affect and mental state analysis and leads to smaller training times in comparison to state of the art. Furthermore, we introduced a novel relational regression loss that aims at learning from the label relationships of the samples during training. The proposed loss uses the distance between label vectors to learn intra-batch latent representation similarities in a supervised manner. We showed that the improved latent representations obtained with the addition of the relational regression loss lead to improved regression output, without the use of large datasets. Finally, we demonstrated the effectiveness of the proposed method on three datasets for schizophrenia symptom severity estimation and for continuous affect estimation, and we showed that our method achieves results comparable to the state-of-the-art -- specifically for symptom severity estimation of schizophrenia, our methodology outperforms the previous state-of-the-art on all scales and symptoms tested and achieves a Pearson's Correlation Coefficient similar to that of human experts.


\begin{acks}
This work is supported by EPSRC DTP studentship (No. EP/R513106/1) and EU H2020 AI4Media (No. 951911).
\end{acks}

\bibliographystyle{ACM-Reference-Format}









\end{document}